\documentclass[conference,compsoc]{IEEEtran}
\usepackage[utf8]{inputenc}
\usepackage{authblk}
\usepackage{setspace}
\usepackage{amsmath,amssymb}
\usepackage{array}
\usepackage{longtable}
\usepackage{tabulary}
\usepackage{graphicx}
\usepackage[T1]{fontenc}
\usepackage{multirow}
\usepackage{makecell}
\usepackage[numbers,sort&compress,square]{natbib}
\newtheorem{definition}{Definition}

\title{SoK: Adversarial Machine Learning Attacks and Defences in Multi-Agent Reinforcement Learning}
\author[1,2]{Maxwell Standen}
\author[1]{Junae Kim}
\author[2]{Claudia Szabo}
\affil[1]{Defence Science and Technology Group}
\affil[2]{The University of Adelaide}

\date{\today}

\begin{document}

\maketitle

\begin{abstract}
    Multi-Agent Reinforcement Learning (MARL) is vulnerable to Adversarial Machine Learning (AML) attacks and needs adequate defences before it can be used in real world applications. 
    We have conducted a survey into the use of execution-time AML attacks against MARL and the defences against those attacks. We surveyed related work in the application of AML in Deep Reinforcement Learning (DRL) and Multi-Agent Learning (MAL) to inform our analysis of AML for MARL.
    We propose a novel perspective to understand the manner of perpetrating an AML attack, by defining Attack Vectors.
    We develop two new frameworks to address a gap in current modelling frameworks, focusing on the means and tempo of an AML attack against MARL, and identify knowledge gaps and future avenues of research.
\end{abstract}


\section{Introduction}
Deep Reinforcement Learning (DRL) has made significant progress over the past decade, from its start playing Atari games \cite{mnih_playing_2013}, to beating humans in board \cite{silver_mastering_2016} and video games \cite{openai_dota_2019, vinyals_grandmaster_2019}, to now addressing important complex safety-critical challenges such as defending computer systems \cite{standen_cyborg_2021}, managing power networks \cite{marot_l2rpn_2020} and driving vehicles \cite{perez-gil_deep_2022}. This increase in complexity has transitioned from fully-observable single-agent environments to partially-observable multi-agent environments. However, to realise positive societal impacts, DRL must overcome the major vulnerability in all deep learning systems to Adversarial Machine Learning (AML) attacks \cite{szegedy_intriguing_2014}, which exploit the neural networks that are the fundamental component of deep learning. 

AML identifies vulnerabilities in machine learning algorithms, such as slight human-imperceptible changes to inputs returning significantly different outputs from neural networks \cite{szegedy_intriguing_2014}. Neural networks are the core of all deep learning, and may generalise to unseen inputs which is key to DRL \cite{mnih_playing_2013}. Despite existing research interest \cite{ilahi_challenges_2021}, there are many challenges with the application of AML techniques to Multi-Agent Reinforcement Learning (MARL), Multi-Agent Learning (MAL), and DRL, including discovering effective attacks \cite{lin_tactics_2017,huang_adversarial_2017,sun_stealthy_2020, qiaoben_strategically-timed_2021} and finding defences that can generalise to unseen attacks \cite{kos_delving_2017,bai_recent_2021}.

AML attacks exist in a large potential solution space. Attacks against supervised image classifiers use approximation techniques such as Fast Gradient Sign Method (FGSM) \cite{goodfellow_explaining_2015} to find adversarial examples for single static images. DRL involves many sequential observations and finding an effective attack requires knowing when, how, and what to attack. Huang et al. \cite{huang_adversarial_2017} showed that approaches such as FGSM attacks were able to attack DRL by altering the observation at every time step. Subsequent research has found attacks that reduced the number of perturbed time-steps required to degrade the agent performance  \cite{lin_tactics_2017,sun_stealthy_2020,qiaoben_strategically-timed_2021}, demonstrating the importance of discovering the best attack timing. 
Adversarial training may defend against AML attacks on supervised learning and DRL \cite{bai_recent_2021}. Adversarial training uses both original and adversarially produced examples to retrain a vulnerable algorithm. In supervised learning, the retrained algorithm is then more robust against similar future attacks \cite{madry_towards_2018}. However, adversarial training in DRL produces a narrow robustness only against the specific attack being used and thus is unable to generalise to other attacks \cite{kos_delving_2017}. 

There is a critical need for MARL practitioners to understand the AML attacks and defences \cite{siva_kumar_adversarial_2020}. A number of previous works have surveyed the state of AML as applied to DRL \cite{ilahi_challenges_2021, chen_adversarial_2019, prorok_beyond_2021, bai_recent_2021, eigner_towards_2021}. However, to the best of our knowledge, there is no work concerning the application of AML against MARL. In this work we address the use of AML techniques on MARL, by employing a new attack perspective called Attack Vector and proposing classifications for AML attacks and defences. Our taxonomy of AML attacks covers how an attack may be deployed, what information it uses, and the attack goal. Our taxonomy of AML defences covers a range of different types of defences, the Attack vectors a defence can counter, when a defence is deployed, and what information a defence requires about an attack. The contributions we make in this work are fivefold:

\begin{itemize}
    \item A survey of AML attacks and defences as applied to MARL, DRL, and MAL more broadly.
    \item A cyber-security inspired perspective on AML attacks against MARL, Attack Vectors.
    \item An improved categorisation on AML defences, which better categorises the different defences against execution-time attacks against MARL.
    \item Two new frameworks for modelling AML attacks against DRL
    that describe combinations of Attack Vectors and attack magnitude,  tempo and location.
    \item An in-depth discussion of future research directions.
\end{itemize} 

\section{Background and Related Work}

We provide a background to MARL. We discuss previous surveys that have covered AML attacks against MARL, DRL, and MAL and existing gaps that we aim to cover.

\subsection{Multi-Agent Reinforcement Learning}
MARL extends DRL to a multi-agent context. Multi-agent environments can be competitive, cooperative, or mixed based on the rewards provided to the agents \cite{busoniu_comprehensive_2008}. \textit{Competitive environments} include but are not limited to zero-sum rewards, where an increase in one agent's reward directly reduces another agent's reward. \textit{Cooperative environments} include rewards that are completely shared by all agents. \textit{Mixed environments} may feature a combination of competitive and cooperative elements in their reward structure. An example of a mixed environment is where there are two competing teams, with agents on the same team receiving the same reward and agents on opposing teams vying for a zero-sum reward.

Two important characteristics of MARL algorithms focus on where the algorithms are trained and executed. \textit{Centralised training} occurs when all agents are trained on a single system and may exploit this by sharing parameters between agents \cite{gupta_cooperative_2017} or using a single critic for all agents \cite{foerster_counterfactual_2018}. \textit{Decentralised training} requires agents to coordinate their training and includes examples such as Independent Q-Learning (IQL) \cite{tan_multi-agent_1993} and Fully Decentralized MARL with Networked Agents \cite{zhang_fully_2018}. A centrally executed MARL algorithm operates similar to a single agent algorithm. However, the structure of the centralised multi-agent system may have greater scalability but worse coordination than a single-agent. This trade-off  occurs because the centralised multi-agent system reduces the coupling between its components. \textit{Decentralised execution} occurs when each agent autonomously and independently executes its policy.

Decentralised MARL \cite{tan_multi-agent_1993, matignon_independent_2012, zhang_fully_2018} faces challenges due to the requirement of stationarity for theoretically guaranteed convergence \cite{matignon_independent_2012}. However, breaking the stationarity requirement does not always lead to failure as demonstrated by IQL \cite{tan_multi-agent_1993}. Each agent in the IQL performs its own Q-learning independent of all other agents. Updating individual agent policies in this manner causes the environment to appear non-stationary from each agent's perspective. However in practice this has been shown to be an effective MARL algorithm in a range of environments \cite{matignon_independent_2012}.

While centralised training with centralised execution is very similar to single-agent DRL, existing hierarchical approaches have shown scalability to larger action spaces \cite{liu_discrete-valued_2021, sukhbaatar_learning_2016}. CommNet \cite{sukhbaatar_learning_2016} is an example of a centrally-trained centrally-executed MARL algorithm. CommNet shares aggregations of the intermediate outputs between its agents, improving agent coordination. However, CommNet requires centralised execution because of the amount of data being shared and the frequency of inter-agent communication. 

There are many approaches to centralised training decentralised execution \cite{blumenkamp_emergence_2021,das_tarmac_2019,foerster_learning_2016-1,foerster_counterfactual_2018,gupta_cooperative_2017,iqbal_actor-attention-critic_2019,lowe_multi-agent_2017,singh_learning_2018}. Differentiable Inter-Agent Learning (DIAL) \cite{foerster_learning_2016-1} learns efficient inter-agent communications. Inter-agent communication allows agents to better coordinate their actions, and is  critical in partially-observable environments as it allows agents to share information. DIAL learns a communication protocol by directly connecting agents through explicit communication channels and allowing the back-propagation of learning gradients across them.  This training requires centralisation, however, it allows the algorithm to find effective communication protocols quicker than random search. To enable decentralised execution, additional noise is added to  communication channels during  training, to represent the discretisation loss present during execution. Then, during execution, the direct connections between agents are replaced with a discretisation/regularisation unit that converts between the continuous outputs of the agents and discrete messages.


\subsection{Related Work}
There are multiple surveys that focus on AML for DRL \cite{ilahi_challenges_2021, bai_recent_2021, chen_adversarial_2019, ren_adversarial_2020}. However our analysis has found that these surveys have limited coverage of AML as applied to MARL and MAL \cite{ilahi_challenges_2021}. Several surveys focus strongly  on Observation Perturbation attacks   \cite{ilahi_challenges_2021, chen_adversarial_2019, ren_adversarial_2020}, including on the information used to craft an attack and when an attack should occur. In our analysis of  previous work that classifies AML attacks  \cite{ilahi_challenges_2021, bai_recent_2021} specifically against MARL, we found no surveys that covered AML defences for MARL; we have identified several that covered AML defences for DRL \cite{ilahi_challenges_2021, chen_adversarial_2019, bai_recent_2021, ren_adversarial_2020}. Our work is unique due to its focus on AML execution-time attacks and defences for DRL, MAL and MARL.

MAL is a major focus of our work and attacks against it can be highly effective as discussed by Ilahi et al. \cite{ilahi_challenges_2021}. Their survey discusses an Observation Perturbation attack against c-MARL by Lin et al. \cite{lin_robustness_2020} and in competitive environments by Gleave et al. \cite{gleave_adversarial_2020}. To build from this work, we have found additional work that investigates attacks against cooperative MARL \cite{lin_robustness_2020, xue_mis-spoke_2022} and cooperative multi-agent supervised learning \cite{tu_adversarial_2021}. We also found a number of AML attacks in competitive MARL \cite{fujimoto_adversarial_2021,uesato_adversarial_2018,guo_adversarial_2021,pan_improving_2021,phan_learning_2020}, which is often overlooked due to the focus of AML on direct Observation Perturbations. 

Observation Perturbation attacks against DRL algorithms have been covered by a number of surveys \cite{ilahi_challenges_2021, chen_adversarial_2019, ren_adversarial_2020}. These have largely focused on the work inspired by Huang et al. \cite{huang_adversarial_2017}. Discovering effective Observation Perturbation attacks with white and black-box information was covered by Chen et al. \cite{chen_adversarial_2019}. Discovering when to attack is a unique problem for AML attacks against DRL and approaches to this question were covered by Ilahi et al. \cite{ilahi_challenges_2021}. We believe our work presents the most extensive survey of Observation Perturbation attacks against DRL to date.


DRL practitioners must understand potential AML attacks and researchers assist in this requirement through the classification of AML attacks such as those proposed in previous work \cite{ilahi_challenges_2021, bai_recent_2021}. Ilahi et al. \cite{ilahi_challenges_2021} presents a taxonomy based on Markov Decision Processes (MDPs) to categorise an attack, which uses the major categories of reward space, action space, state space, and agent space to cover both test-time and training-time attacks. Bai et al. \cite{bai_recent_2021} classified attacks based on the information an adversary requires to create the attack. However, neither of these classifications support the categorisation of AML attacks unique against MARL, such as attacks against communications in cooperative MARL and adversarial policies in competitive MARL. To address this gap, we have developed a classification of AML attacks against MARL and DRL that covers the means of attack, the aim of the attack, and the knowledge of the target required by an adversary.
 


No AML defences for MARL have been covered by previous surveys to the best of our knowledge. To find related work, we consider surveys that cover AML defences for DRL \cite{ilahi_challenges_2021, chen_adversarial_2019, bai_recent_2021, ren_adversarial_2020}. Ilahi et al. \cite{ilahi_challenges_2021} propose a taxonomy that features a number of classifications including adversarial training, defensive distillation, robust learning, and adversarial detection, however it fails to consider Competitive Training, which aims to use a competitor in the environment to train the DRL agent. Our work addresses this gap. Chen et al. \cite{chen_adversarial_2019} also use Input Alteration, which includes the minor categories of various adversarial training, as the alteration covers both test and training time. They further propose two other categories of altering objective function and altering network structure to improve the robustness against AML attacks. However, the examples they use for AML defence are from supervised learning. Bai et al. \cite{bai_recent_2021} focussed on adversarial training which was a category in both Ilahi et al. \cite{ilahi_challenges_2021} and Chen et al. \cite{chen_adversarial_2019}, however Bai et al. restructured adversarial training as a competitive multi-agent problem. This move towards using multi-agent perspectives to better understand AML attacks is one that we expand upon. Adversarial Training, Input Alteration, and Robust Learning were categories used by Ren et al. \cite{ren_adversarial_2020}. We have drawn from these existing classifications and our own analyses and present a classification system for defences for both MARL and DRL.

Our paper is unique in focusing on execution-time AML attacks against DRL and defences against those attacks. Ilahi et al. \cite{ilahi_challenges_2021} covered both execution-time and training-time attacks and defences. Chen et al. \cite{chen_adversarial_2019} focused on execution-time attacks against DRL but looked at AML defences for supervised learning. Bai et al. \cite{bai_recent_2021} focused on adversarial training as a defence against execution-time attacks in both DRL and supervised learning. Ren et al. \cite{ren_adversarial_2020} covered AML attacks and defences against the whole deep learning field including supervised learning and DRL. Our unique focus allows us to better analyse the problem of defending a DRL algorithm from execution-time attacks. Related work has covered aspects of AML applied to DRL, however there remain gaps in the coverage of existing surveys around AML attacks and defences for MARL, and the classification of AML attacks and defence when applied to DRL and MAL. 


\section{Methodology}
We used a literature review methodology based on \cite{kitchenham_systematic_2009}. We first identified and refined a number of research questions. From these questions, we identified keywords and constructed search strings. The search strings were used to find relevant papers in top-tier conferences that publish DRL, AML and MAL papers.  We augmented this collection by forward snowballing \cite{wohlin_guidelines_2014} papers on the topic to discover very recent breakthroughs in the topic, for a total of 350 papers. From this collection, we rejected papers that did not match our accept/reject criteria, resulting in 85 papers. Finally we extracted data from these papers using a set of guiding questions, resulting in 57 papers in our final set. 


\subsection{Research Questions and Search Strings}
For a broad overview of AML attacks against MARL, MAL, and DRL approaches, our research questions are:
\begin{itemize}
    \item RQ1: What AML attacks exploit vulnerabilities in MARL, MAL, and DRL during execution?
    \item RQ2: What AML defences mitigate execution-time attacks against MARL, MAL and DRL? 
\end{itemize}






We used two search strings to find papers that would address our research questions.
The first string  is {\ttfamily "reinforcement AND (training OR learning) AND (adversarial OR attack)} and found 160  papers that relate to adversarial attacks and defences against DRL and MARL.
The second string  is {\ttfamily (multiagent OR multi-agent OR (multi AND agent)) AND (policy OR learning) AND (adversarial OR attack)} and found 88 papers that relate to adversarial attacks and defences against MAL and MARL. 

We limited our search to high-quality conferences that published work in DRL, MAL, and MARL since 2010. Our search strings were used in Scopus and external proceedings servers where necessary to search the following conferences:
\begin{itemize}
    \item Intl. Joint Conference on Autonomous Agents and Multiagent Systems 
    \item Intl. Conference on Machine Learning 
    \item Usenix Security Symposium 
    \item Intl. Joint Conference on Artificial Intelligence 
    \item Intl. Conference on Learning Representations 
    \item National Conference of the American Association for Artificial Intelligence 
    \item IEEE Symposium on Security and Privacy 
    \item ACM Conference on Computer and Communications Security 
    \item ACM International Conference on Knowledge Discovery and Data Mining
    \item Advances in Neural Information Processing Systems
\end{itemize}
This list does not cover 2010 to 2021 for the relevant conferences, as some  did not run in some years.

Finally, we used several well known papers that address our research questions \cite{tu_adversarial_2021,sun_stealthy_2020,blumenkamp_emergence_2021} and used a forward snowball search method \cite{wohlin_guidelines_2014} to find 147 papers that cited them. These papers were chosen because they made significant discoveries that are relevant to our search. 


We identified accept and reject criteria and applied these to the papers in our set. 
For a paper to be excluded, it must either fail to meet all of the accept criteria or meet any of the reject criteria. Our accept criteria were:

\begin{itemize}
    \item Uses either an execution-time adversarial attack or a defence against an execution-time adversarial attack
    \item The attack or defence are used on either an offline deep reinforcement learning or a cooperative multi-agent decentrally-executable deep learning algorithm
\end{itemize}

Our reject criteria were:
\begin{itemize}
    \item Number of pages is less than five
    \item Does not feature any experimentation
\end{itemize}

We define adversarial attacks as the intentional manipulation of aspects of an environment to reduce the performance of a target network and cause a target agent to take irrational or self-defeating actions. We further limit the scope of our study to execution-time attacks.


\section{AML Attacks}
\begin{figure}
    \centering
    \includegraphics[width=\columnwidth]{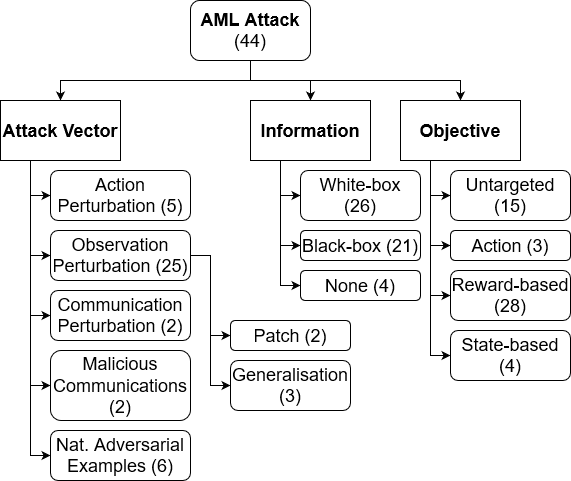}
    \caption{Classification of Adversarial Machine Learning (AML) attacks against Deep Reinforcement Learning (DRL) and the number of papers that considered those categories. }
    \label{fig:att_class}
\end{figure}

Our analysis of the papers in our set allows us to propose a new classification system for discussing AML attacks, shown in Figure \ref{fig:att_class}. We have identified three important properties of an AML attack, namely, \textit{Attack Vector}, \textit{Information}, and \textit{Objective}. 
We consider the adversary's knowledge capability by classifying what \textit{information} the adversary knows about the target algorithm. 
We use the standard categories of white-box information \cite{zheng_vulnerability_2021,behzadan_whatever_2017,sun_who_2021,lin_robustness_2020,wang_adversarial_2020,weng_toward_2020,fujimoto_adversarial_2021,chan_adversarial_2020,tu_adversarial_2021,huang_adversarial_2017,wu_adversarial_2021,hussenot_copycat_2020,praveen_kumar_critical_2021,kos_delving_2017,pan_improving_2021,huai_malicious_2020,tekgul_real-time_2022,zhang_robust_2020-1,tretschk_sequential_2018,lee_spatiotemporally_2020,sun_stealthy_2020,qiaoben_strategically-timed_2021,lin_tactics_2017,buddareddygari_targeted_2021,qiaoben_understanding_2021}, black-box information \cite{blumenkamp_emergence_2021,guo_adversarial_2021,inkawhich_snooping_2020,chen_understanding_2021,behzadan_vulnerability_2017,wang_adversarial_2020,gleave_adversarial_2020,chan_adversarial_2020,tu_adversarial_2021,huang_adversarial_2017,qu_attacking_2021,hussenot_copycat_2020,guo_edge_2021,pan_improving_2021,garcia_learning_2020,phan_learning_2020,xue_mis-spoke_2022,russo_optimal_2019,lee_query-based_2021,uesato_rigorous_2019,pattanaik_robust_2018} as presented by Bai et al. \cite{bai_recent_2021}, with the additional category for attacks that use no information \cite{tessler_action_2019,korkmaz_daylight_2020,korkmaz_assessing_2021,korkmaz_adversarial_2021}.

The final category is the \textit{Objective} of the attack. A common classification are untargeted attacks \cite{inkawhich_snooping_2020,behzadan_vulnerability_2017,behzadan_whatever_2017,wang_adversarial_2020,korkmaz_daylight_2020,fujimoto_adversarial_2021,tu_adversarial_2021,wu_adversarial_2021,korkmaz_adversarial_2021,korkmaz_assessing_2021,kos_delving_2017,guo_edge_2021,tekgul_real-time_2022,pattanaik_robust_2018,qiaoben_understanding_2021}, which do not pursue a specific goal but instead aim to alter the behaviour of an agent. Untargeted misclassification was originally identified in a taxonomy for AML attacks on supervised learning algorithms \cite{papernot_limitations_2016}, which also proposed the objective of targeted misclassification. In a DRL context, targeted misclassification means causing the agent to select a specific action \cite{inkawhich_snooping_2020,hussenot_copycat_2020,qiaoben_understanding_2021}. Objectives unique to reinforcement learning methods are \textit{state-based} and \textit{reward-based targets}. Attack that use state-based objectives aim to influence the agent to move towards a specific goal state \cite{lin_tactics_2017,buddareddygari_targeted_2021,chen_understanding_2021,uesato_rigorous_2019}. Reward-based objectives focus on altering the reward returned by the environment during execution and include goals such as maximising the adversary's reward and minimising the rewards of the targeted agent \cite{blumenkamp_emergence_2021,guo_adversarial_2021,inkawhich_snooping_2020,zheng_vulnerability_2021,sun_who_2021,lin_robustness_2020,gleave_adversarial_2020,weng_toward_2020,tessler_action_2019,fujimoto_adversarial_2021,chan_adversarial_2020,huang_adversarial_2017,qu_attacking_2021,praveen_kumar_critical_2021,pan_improving_2021,garcia_learning_2020,phan_learning_2020,huai_malicious_2020,xue_mis-spoke_2022,russo_optimal_2019,lee_query-based_2021,zhang_robust_2020-1,tretschk_sequential_2018,lee_spatiotemporally_2020,sun_stealthy_2020,qiaoben_strategically-timed_2021,lin_tactics_2017}.



\section{AML Attack Vectors}
\label{sec:aml_att_vec}

We have identified five major Attack Vectors that exploit different aspects of MARL, DRL, or MAL during execution-time, namely, Action Perturbation, Observation Perturbation, Communication Perturbation, Malicious Communications, and Natural Adversarial Examples. Adversaries use \textit{Action Perturbations} against DRL and MARL to alter the action selection of an agent, which affects the state transition function and the observation function. \textit{Observation Perturbations} allows adversaries to directly inject data into an agent, altering the action selected by an agent to induce an adversarial effect. \textit{Communication Perturbation} intercepts and alters agent communications to attack MAL and MARL. \textit{Malicious Communications} provides adversarial agents with a communication capability that exploits the use of communications without intercepting or altering existing communications in MARL. \textit{Natural Adversarial Examples} use the environmental actions of an adversary to alter the state of the environment and indirectly attack a MARL algorithm.

\subsection{Action Perturbation}

\begin{table*}[ht]
\centering
\caption{Action Perturbation attacks}
\begin{tabulary}{1.0\textwidth}{LLLLLLLLL} 
\hline
\textbf{Attack Name}                            & \textbf{Information for Attack} & \textbf{Objective of Attack}   & \textbf{Attacked Algorithm(s)} & \textbf{Framework} & \textbf{Action Space} & \textbf{Experimental Platform}  \\ 
\hline
\textbf{Targeted Adversarial Perturbations} \cite{lee_query-based_2021}       & Black-box                       & Reward-based & PPO                         & MDP            & Continuous            & Point-Goal, Car-Goal            \\ 
\textbf{\textbf{Myopic Action Space (MAS) attack}} \cite{lee_spatiotemporally_2020} & White-box                       & Reward-based         & DDQN, PPO                   & MDP            & Continuous            & MuJoCo, OpenAI Gym              \\ 
\textbf{Look-ahead Action Space (LAS) attack} \cite{lee_spatiotemporally_2020}     & White-box                       & Reward-based         & DDQN, PPO                   & MDP            & Continuous            & MuJoCo, OpenAI Gym              \\ 
\textbf{Probabilistic Action Perturbation} \cite{tessler_action_2019}        & None                            & Reward-based         & DDPG                        & PR-MDP        & Continuous            & MuJoCo                          \\ 
\textbf{\textbf{Noisy Action Perturbation}} \cite{tessler_action_2019}       & None                            & Reward-based         & DDPG                        & NR-MDP         & Continuous            & MuJoCo                          \\ 
\textbf{EDGE} \cite{guo_edge_2021}                                     & Black-box                       & Untargeted                  & EDGE                        & MDP            & Discrete, Continuous  & ALE, MuJoCo                     \\
\textbf{Model-based Attack \cite{weng_toward_2020}} & White-box & Reward-based & D4PG & MDP & Continuous & MuJoCo \\
\hline
\end{tabulary}
\label{tab:action_attacks}
\end{table*}

Action Perturbations are a form of attack unique to DRL and MARL in which the adversary directly alters the action of an agent. Altering this action reduces the performance of the target agent because the new action will be the worst possible action for that agent based on an adversary's capability. Additionally, this alteration may cause the environment to transition to a state to which the target agent is unable to optimally act.

A number of papers consider Action Perturbations \cite{lee_query-based_2021, lee_spatiotemporally_2020, tessler_action_2019, guo_edge_2021, weng_toward_2020}. The techniques covered by these papers are presented in Table \ref{tab:action_attacks}. These techniques are able to use a range of information from white-box \cite{lee_spatiotemporally_2020, weng_toward_2020}, black-box \cite{lee_query-based_2021, guo_edge_2021}, and no information \cite{tessler_action_2019}. These techniques generally used a reward-based objective \cite{lee_spatiotemporally_2020,lee_query-based_2021,tessler_action_2019, weng_toward_2020} with the exception of an attack using EDGE \cite{guo_edge_2021} which was untargeted. Action Perturbation attacks were able to target the state-of-the-art Proximal Policy Optimisation (PPO) algorithm \cite{schulman_proximal_2017} as well as the commonly used Double Deep Q-Network (DDQN) \cite{van_hasselt_deep_2018} and Deep Deterministic Policy Gradient (DDPG) \cite{lillicrap_continuous_2019} algorithms. MDPs were used as the framework for all of the attacks except for those by Tessler et al. \cite{tessler_action_2019}, which used MDP variants that modelled Action Perturbation attacks. All of the attacks were demonstrated against continuous action spaces, while EDGE was also shown to be effective against discrete action spaces.

For attacks on continuous action spaces \cite{lee_query-based_2021, lee_spatiotemporally_2020, tessler_action_2019, guo_edge_2021, weng_toward_2020}, the perturbation alters the action selection based on the attack magnitude. Action Perturbations may reveal vulnerabilities in the uncertain execution of actions. This risk of uncertain action execution is common to the robotics domain applications, thus the simulation environment MuJoCo \cite{todorov_mujoco_2012} was used to demonstrate the applicability of the attack \cite{lee_spatiotemporally_2020, tessler_action_2019, guo_edge_2021, weng_toward_2020}. Furthermore, an adversary may be able to use a cyber-physical attack to influence the execution of actions and achieve an Action Perturbation attack.

Action Perturbation attacks may also reveal weaknesses in the generalisation of an algorithm to states in the environment, which is similar to Natural Adversarial Examples. Both of these vectors aim to move the environment to a state to which the target algorithm non-optimally responds. This vulnerability has explicitly been explored in both discrete \cite{guo_edge_2021} and continuous action spaces \cite{tessler_action_2019}. 

The vulnerability an Action Perturbation attack exploits cannot easily be identified using the metrics of reward or win rate. However these metrics are commonly used in the evaluation of Action Perturbation attacks \cite{lee_query-based_2021, lee_spatiotemporally_2020, tessler_action_2019, guo_edge_2021, weng_toward_2020}. We believe that a metric such as counterfactual regret \cite{zinkevich_regret_2007} may better evaluate the effectiveness of Action Perturbation attacks. Regret measures the difference between the action an agent takes and the best possible action it could have taken in hindsight. The regret at the point of attack would indicate the level of vulnerability an algorithm has to a worst case action and the regret in the steps following an attack would indicate the degree of the generalisation vulnerability.




\subsection{Observation Perturbations}

Observation Perturbation attacks, also known as Evasion attacks \cite{tabassi_taxonomy_2019} in supervised learning, expose a vulnerability in an agent that may be exploited by an adversary with the capability of altering the observation received by an agent. This capability requires the adversary to alter the environment \cite{brown_adversarial_2017} or an agent's sensors through a cyber attack \cite{ferdowsi_robust_2018}. The adversary induces the agent to take an action that may differ from the original action the agent would have taken based on the unaltered observation. Thus, Observation Perturbations may be a path to achieving Action Perturbation attacks.

We present many of these Observation Perturbation attacks in Table \ref{tab:evasion_att} in the Appendix. This table covers the information used by the attacks and the objective of the attacks. We also cover additional information such as if the attack is a transfer attack \cite{demontis_why_2019}, which algorithms the attack has been demonstrated against, and the framework that the original paper used when presenting the attack.

All of the Observation Perturbation attacks that we found altered continuous observation spaces. Many of these attacks were against unstructured images, however an attack was presented that targeted the structured observations of an energy management system of an electric vehicle \cite{wang_adversarial_2020}. The data collected from the vehicle was presented to the agent as a vector of continuous values, which was smaller and contained a greater density of information than images. The empirical results showed the effectiveness of the attack, which suggests that attacks against other structured observations, such as those used in autonomous cyber defence \cite{standen_cyborg_2021}, may also be effective. However, the observations in an autonomous cyber defence environment may be discrete and it is unclear how effective these attacks would be against discrete observation spaces.

We have identified two unique sub-categories of Observation Perturbation attacks based on the magnitude and localisation of an attack. 
\textit{Patch attacks} target a small section of the observation but may use a larger magnitude. \textit{Generalisation attacks} use both a large magnitude and may affect the whole image, but are restricted to 'natural' alterations of the observation, such as rotations or colour swaps.

\noindent \textbf{Patch Attacks}
 occur when a small area of an observation is altered. They are easier to realise than other Observation Perturbation attacks, as an adversary only needs to change a small part of the observation. We believe that the localisation of  Patch attacks will reduce their effectiveness compared with other Observation Perturbation attacks of similar magnitude but without the localisation restriction. This comparison has not yet been explored.

Patch attacks have been demonstrated against supervised learning algorithms \cite{brown_adversarial_2017} and we have found two papers that use this attack against DRL \cite{buddareddygari_targeted_2021,huai_malicious_2020}. A Targeted Physical attack \cite{buddareddygari_targeted_2021} controls the appearance of a fixed-size square object that exists in the observation. UniAck \cite{huai_malicious_2020} generates a universal perturbation that attacks an the interpretation module of a DRL algorithm to identify the non-perturbed region of an image as the cause of the action.


\noindent \textbf{Generalisation Attacks}
 use 'natural' alterations to the observation to degrade the performance of an algorithm. These attacks are untargeted and aim to change the observation in a way that a human could easily adapt. These attacks include changing  colours, rotating, introducing compression artefacts, or altering the dimensions of the observation. 

Generalisation attacks vary in their ability to be realised. Rotating a sensor to attack an agent may be easier to achieve than changing the colours that the sensor perceives. Generalisation attacks expose potential weaknesses in a DRL algorithm to basic changes in the environment to which a human would be easily able to adapt. We only found three approaches to generalisation attacks  \cite{korkmaz_daylight_2020,korkmaz_adversarial_2021,korkmaz_assessing_2021}, that all attacked DDQN and used a MDP framework.


\subsection{Communication Perturbation}
Communication Perturbation exploits vulnerabilities in communications between agents. An adversary may exploit this vulnerability if they are capable of altering the messages sent through explicit channels in a cooperative multi-agent system. Implicit communication uses an agent's observation of the environment to communicate information, and so perturbation attacks against implicit communication would be considered a type of Observation Perturbation attack. Observation Perturbation overlaps with Communication Perturbation as an environment may include the message in the observation. Messages sent from other agents may be considered actions and so there is an overlap between Action Perturbation and Communication Perturbation. Separately categorising Communication Perturbation from the other forms of perturbation reveals unique challenges of AML attacks against MARL.

MARL algorithms find communication protocols that use explicit communication in partially-observed cooperative environments. Techniques such as CommNet \cite{sukhbaatar_learning_2016} use the output of a layer of a neural network from one agent as the direct input to a layer in a different agent. Communication Perturbations against this form of communication can be highly effective as shown by Xue et al. \cite{xue_mis-spoke_2022}. Another paradigm of communication embeds the messages as continuously-valued observations, thus AML attacks against this form of communication are similar to those used in Patch and other Observation Perturbation attacks. 

Discretely-valued communications are another form of communication that is common to MARL. RIAL \cite{foerster_learning_2016-1} treats messages as a part of the action-space, whereas DIAL \cite{foerster_learning_2016-1} uses a discretisation/regularisation unit to turn a continuously-valued message into a discrete message. The robustness of discrete messages to AML attack is unclear as all AML attacks that we found attacked continuous observations. 

Two papers consider the Communication Perturbation attack vector \cite{xue_mis-spoke_2022,tu_adversarial_2021}. Xue et al. \cite{xue_mis-spoke_2022} present a reward-based black-box attack against CommNet \cite{sukhbaatar_learning_2016}, TarMAC \cite{das_tarmac_2019}, and NDQ \cite{wang_learning_2020-2}. Tu et al. \cite{tu_adversarial_2021} present the only supervised learning algorithm discovered in our search that fits our criteria. They train a decentralised algorithm that uses emergent communications to recognise an object with information from multiple perspectives. An adversary is then introduced that perturbs the communications to reduce the effectiveness of the object recognition algorithm. Tu et al. \cite{tu_adversarial_2021} demonstrated both a white-box attack and transfer black-box attack against Multi-View ShapeNet.

Communication Perturbation attacks may be more easily realisable than the more commonly explored Observation Perturbation attacks. Cyber security attackers may use a technique called Meddler-In-The-Middle (MITM) \cite{poddebniak_why_2021} which allow them to intercept and alter messages that are sent over a network. The messages sent by MARL agents may be indecipherable to humans, and so the constraints that are normally imposed, such as magnitude of attack or location of attack may not apply. Without these constraints an adversary could completely change the messages sent between agents. This change is similar to Malicious Communication, except that the message is coming from a trusted source. None of the techniques that we found use an unlimited white-box attack against communications, however as communications may make up a small part of an agent's observation, we believe that its effectiveness may be similar to that of a Patch attack. 

\subsection{Malicious Communications}

Malicious Communications exploit the vulnerability that occurs when an adversary is able to send messages to a target agent. Adversarial messages are used by both Malicious Communications and Communication Perturbations. However, Malicious Communication attacks craft a new message instead of altering an existing message. An adversary may realise Malicious Communications more easily than Communication Perturbation, as the adversary does not need to intercept and alter an existing message. We believe that defences against Malicious Communications may be more effective than against Communication Perturbations, as  messages do not originate from a trusted source.

Exploration of Malicious Communications in the literature has been limited \cite{blumenkamp_emergence_2021, qu_attacking_2021}. The attacks presented in these papers are limited to using black-box information, however we believe that white-box information could also be used to craft more devastating and effective attacks. Blumenkamp et al. \cite{blumenkamp_emergence_2021} trained then fixed the policy of a MARL system and then trained an adversarial agent, with Malicious Communications capabilities against that system. However, the primary goal of the adversarial agent was to maximise its own reward, with the reduction in performance of the cooperative system being a secondary effect. Qu et al. \cite{qu_attacking_2021} also took the approach of maximising the adversary's reward. Their adversary attacked a system of decentralised sensors with a central agent receiving data from those sensors. The cooperative sensors used a fixed policy to determine what information to communicate.


The communication paradigm employed should influence the effectiveness of Malicious Communications. Broadcast is the most common communication paradigm in MARL \cite{zhu_survey_2022}, in which all agents send a message that is received by all other agents in the system. Neither of the papers we found used broadcast communications. Blumenkamp et al. \cite{blumenkamp_emergence_2021} aggregated received messages and propagated this to nearby agents, such that those agents that were closer had a greater effect on the message that was received. Qu et al. \cite{qu_attacking_2021} used a many-to-one communication where the many sensors were communicating back to the central agent. We believe that a broadcast malicious message should be highly effective because it is able to simultaneously affect all agents in the system.


\subsection{Natural Adversarial Examples}

Natural Adversarial Examples exploit an agent vulnerability by finding naturally occurring environment states and observations that have an adversarial effect on agents. This type of attack is used in competitive MARL to exploit weaknesses in an opponent's policy. Both Malicious Communication and Natural Adversarial Examples use valid actions in the environment to attack a target, however, Natural Adversarial Examples may only indirectly affect a target.

This type of attack is the most feasible of all the Attack Vectors considered, however it is also both the most difficult and least effective  attack. The only requirement for an adversary is that it can take actions in an environment. However, this attack may be countered by both AML defences and improvements to the DRL agent generalisation.

An adversary must find states and observations that exist in the environment that have either not been encountered or generalised by the target policy. Several papers we found consider Natural Adversarial Examples \cite{uesato_adversarial_2018, wu_adversarial_2021, guo_adversarial_2021, pan_improving_2021, phan_learning_2020, gleave_adversarial_2020} and their attacks are presented in Table \ref{tab:nat_adv_ex} along with the information required for the attack, the objective of the attack, if a transfer attack is used, the algorithms that were attacked, and the framework used in the paper to present the attack.


A major type of Natural Adversarial Example attack was pioneered by Gleave et al. \cite{gleave_adversarial_2020} and extended by Guo et al. \cite{guo_adversarial_2021}. These attacks used competitive MARL against a fixed target policy to find the actions they could perform in the environment that cause the opponent to lose. Gleave et al. \cite{gleave_adversarial_2020} use a zero-sum reward to minimise the target's reward, and found effective strategies in MuJoCo. The extension by Guo et al. \cite{guo_adversarial_2021}, that relaxed the zero-sum constraint, was highly effective in Starcraft II environments.


\begin{table*}[ht]
\centering
\caption{Natural Adversarial Example Attacks}
\begin{tabulary}{1.0\textwidth}{p{4cm}p{2.2cm}p{2.5cm}p{1.5cm}p{3cm}l} 
\hline
\textbf{Name of attack}              & \textbf{Information for attack} & \makecell[lt]{\textbf{Objective} \\ \textbf{of attack}} & \makecell[lt]{\textbf{Transfer} \\ \textbf{attack}} & \textbf{Attacked algorithms}                                                                         & \textbf{Framework}   \\ 
\hline
\textbf{Failure Search \cite{uesato_adversarial_2018}}              & Black-box                       & State-based      & $\times$                       & D4PG & MDP              \\ 
\textbf{Adversarial Policy \cite{wu_adversarial_2021}} & White-box                       & Untargeted       & $\times$                       & PPO                                                           & MDP          \\ 
\textbf{Adversarial Policy \cite{gleave_adversarial_2020, guo_adversarial_2021}}          & Black-box                       & Reward-based       & $\times$                       & PPO                                                                                                & SG  \\ 
\textbf{White-box Adversary \cite{pan_improving_2021}}               & White-box           & Reward-based       & $\times$                  & KAIST, PARL, NANYANG, D3QN                                                                         & MDP              \\ 
\textbf{Black-box Adversary \cite{pan_improving_2021}}               & Black-box             & Reward-based       & $\checkmark$                   & KAIST, PARL, NANYANG, D3QN                                                                         & Failure-MDP              \\ 
\textbf{Antagonists \cite{phan_learning_2020}}                 & Black-box                       & Reward-based       & $\times$                       & DQN, QMIX                                                                                          & POSG             \\ 
\hline
\end{tabulary}
\label{tab:nat_adv_ex}
\end{table*}

\section{AML Defences}
\begin{figure}
    \centering
    \includegraphics[width=\columnwidth]{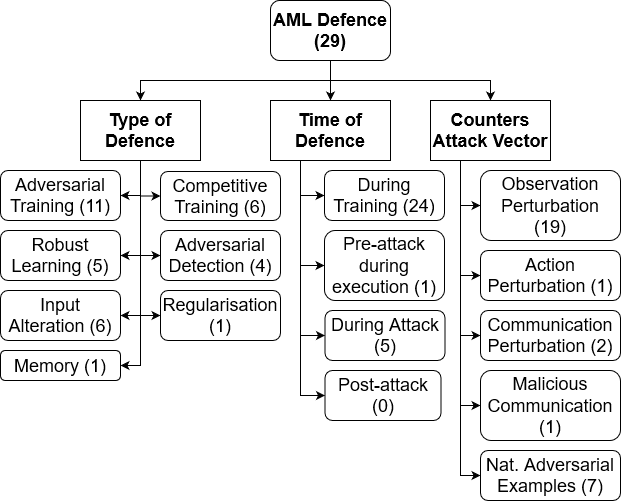}
    \caption{Classification of AML defences for DRL and the number of papers that considered those categories.} 
    \label{fig:def_class}
\end{figure}

We consider several categories in the classification of AML defences for MARL, DRL and MAL, namely, the type of defence, when the defence occurs, and what attacks the defence counters. 
We have identified several types of AML defences that are used to defend MARL and DRL algorithms from AML attacks. These are Adversarial Training, Competitive Training, Robust Learning, Adversarial Detection, Input Alteration, Memory, and Regularisation.
When considering when a defence is applied, we identify four general times, namely, during training, during execution before an attack, during an attack, and after an attack. Figure \ref{fig:def_class} shows our classification.

\noindent \textbf{Adversarial Training}
\cite{bai_recent_2021} retrains the original agent against the AML attack. Adversarial training occurs during the training phase of the ML pipeline, and its main purpose is to counter Observation Perturbations. It has also been shown to be effective in countering Communication Perturbations \cite{tu_adversarial_2021}. Table \ref{tab:adv_training} presents the adversarial training techniques and the attack vector countered, the defence's impact on  performance, the algorithms being defended, and the framework the paper used to present the defence.

We exclude online training from this category despite its potential for performing adversarial training during execution. Online training against an adversary poses a risk as the adversary may influence the data collected from the environment. This adversary-influenced data will be used by the algorithm during the online training process. An adversary may intentionally poison this training data to cause the algorithm to learn a poor policy \cite{kiourti_trojdrl_2020, han_adversarial_2020}. Thus online adversarial training  needs to handle data poisoning attacks before it can be deployed as an effective defence.

\begin{table}[ht]
\centering
\caption{Adversarial Training Defences}
\resizebox{\columnwidth}{!}
{
\begin{tabular}{ p{2.0cm} p{2.0cm} p{1.7cm} p{1.3cm} l } 
\hline
\makecell[lt]{\textbf{Name of} \\ \textbf{defence}} & \makecell[lt]{\textbf{Counters} \\ \textbf{attacks}}  & \makecell[lt]{\textbf{Impact on} \\ \textbf{performance}} & \makecell[lt]{\textbf{Defended} \\ \textbf{algorithm(s)}} & \textbf{Framework}  \\ 
\hline
\textbf{Adversarial training \cite{korkmaz_adversarial_2021}}                                     & \makecell[lt]{Observation \\ Perturbation}     & Negative                       & DDQN                      & MDP         \\ 
\textbf{Robustifying models \cite{tu_adversarial_2021}}                                      & \makecell[lt]{Communication\\ Perturbation}  & Positive                       & Multi-View ShapeNet        & N/A           \\ 
\textbf{DQWAE \cite{ohashi_deep_2021}} & \makecell[lt]{Observation \\ Perturbation}     & Positive                       & DQN                        & MDP         \\ 
\textbf{Adversarial training \cite{kos_delving_2017}}                    & \makecell[lt]{Observation \\ Perturbation}     & \makecell[lt]{Not \\ evaluated}                  & A3C                        & MDP         \\ 
\textbf{Adversarial training \cite{korkmaz_investigating_2021}}                                     & \makecell[lt]{Observation \\ Perturbation}     & \makecell[lt]{Not \\ evaluated}                  & SA-DQN                     & MDP         \\ 
\textbf{SA DRL \cite{zhang_robust_2020-1}}                                    & \makecell[lt]{Observation \\ Perturbation}     & Positive                       & PPO, DDPG, DQN             & SA-MDP          \\ 
\textbf{RADIAL-RL \cite{oikarinen_robust_2021}}                      & \makecell[lt]{Observation \\ Perturbation}     & Mixed                          & DQN, A3C, PPO              & MDP         \\ 
\makecell[lt]{\textbf{Robust} \\ \textbf{training \cite{pattanaik_robust_2018}}}                                          & \makecell[lt]{Observation \\ Perturbation}     & Positive                       & DDQN, DDPG                 & MDP         \\ 
\textbf{Adversarial training \cite{behzadan_whatever_2017}}                                     & \makecell[lt]{Observation \\ Perturbation}     & \makecell[lt]{Not \\ evaluated}                  & DQN                        & MDP         \\ 
\textbf{PA-ATLA \cite{sun_who_2021}}                                                  & \makecell[lt]{Observation \\ Perturbation}     & Mixed                          & DQN, A2C, PPO              & MDP         \\
\hline
\end{tabular}
}
\label{tab:adv_training}
\end{table}

\noindent \textbf{Competitive Training}
is similar to Adversarial Training, however instead of training the agent against perturbed observations, the target is trained against an opponent agent. This defence is largely deployed against Natural Adversarial Examples but has also been shown to be effective against Action Perturbations and Communication Perturbations. Table \ref{tab:comp_training} presents the defences in this category, which attacks the defence counters, the impact on performance, the algorithm being defended, and the framework used by the paper that presented the defence.

\begin{table}[ht]
\centering
\caption{Competitive Training Defences}
\resizebox{\columnwidth}{!}
{
\begin{tabular}{ p{2.1cm} p{2.0cm} p{1.7cm} p{1.4cm} p{1.5cm} } 
\hline
\makecell[lt]{\textbf{Name of} \\ \textbf{defence}} & \makecell[lt]{\textbf{Counters} \\ \textbf{attacks}}  & \makecell[lt]{\textbf{Impact on} \\ \textbf{performance}} & \makecell[lt]{\textbf{Defended} \\ \textbf{algorithm(s)}} & \textbf{Framework}  \\ 
\hline
\textbf{RARL \cite{pinto_robust_2017}}                   & Natural Adversarial Examples & Positive                       & TRPO                       & SG              \\ 
\makecell[lt]{\textbf{Probabilistic} \\ \textbf{Operator \cite{tessler_action_2019}}} & \makecell[lt]{Action \\ Perturbations}         & \makecell[lt]{Not \\ evaluated}                   & DDPG                       & PR-MDP          \\ 
\makecell[lt]{\textbf{Noisy} \\ \textbf{Operator \cite{tessler_action_2019}}}         & \makecell[lt]{Action \\ Perturbations}         & \makecell[lt]{Not \\ evaluated}                   & DDPG                       & NR-MDP          \\ 
\makecell[lt]{\textbf{Adversary} \\ \textbf{Resistance \cite{guo_adversarial_2021}}}   & Natural Adversarial Examples & \makecell[lt]{Not \\ evaluated}                   & PPO                        & SG              \\ 
\textbf{Adversarial training \cite{pan_improving_2021}}   & Natural Adversarial Examples & Positive                       & KAIST, PARL, NANYANG, D3QN & MDP             \\ 
\textbf{ARTS \cite{phan_learning_2020}}                   & Natural Adversarial Examples & Negative                       & DQN, QMIX                  & POSG            \\ 
\textbf{MACRL \cite{xue_mis-spoke_2022}}                  & Communication Perturbations  & \makecell[lt]{Not \\ evaluated}                  & CommNet, TarMAC, NDQ       & Dec-POMDP-Com   \\
\hline
\end{tabular}
}
\label{tab:comp_training}
\end{table}

\noindent \textbf{Robust Learning}
increases the robustness of an agent in an environment. These methods do not consider specific attacks and instead consider robustness against specific adversary capabilities, such as the magnitude of perturbation. Table \ref{tab:rob_learn} shows robust learning techniques, their impact on performance, and which algorithm is defended.

A limitation of our work is that our data collection only found papers that considered the improved robustness in the context of defending against an AML attack. We believe that the robust learning category could be extended to consider approaches that do not consider specific AML adversary capabilities, but instead look at generally improving the robustness of an agent to other conditions such as robustness against non-stationary environments \cite{lecarpentier_non-stationary_2019}.

\begin{table}[ht]
\centering
\caption{Robust Learning Defences}
\begin{tabular}{lll} 
\hline
\textbf{Name of defence}                        & \makecell[lt]{\textbf{Impact on} \\ \textbf{performance}} & \makecell[lt]{\textbf{Defended} \\ \textbf{algorithm(s)}}  \\ 
\hline
\textbf{A2PD \cite{qu_adversary_2021}} & Positive                       & DQN                         \\ 
\textbf{CARRL \cite{everett_certifiable_2021}}                                  & Not evaluated                  & DQN                         \\ 
\textbf{CROP \cite{wu_crop_2022}}                                   & Not evaluated                  & DQN                         \\ 
\textbf{CPPO \cite{ying_towards_2021}}                                   & Positive                       & PPO                         \\ 
\textbf{TRC \cite{kim_trc_2022}}                                    & Positive                       & CPO, TRPO-L                 \\
\hline
\end{tabular}
\label{tab:rob_learn}
\end{table}

\noindent \textbf{Input Alteration}
mitigates an AML attack by removing the adversarial component of the observation or communication. Table \ref{tab:input_alt} shows the input alteration defences, their requirement for adversarial detection, which attacks they counter, their impact on performance, the name of the algorithm being defended, and the framework used by the paper that presented the defence. 

Methods that do not require the detection of adversarial inputs alter the observation based on assumptions about an adversary's capabilities, such as the magnitude of perturbation. These defences have been considered in supervised learning contexts \cite{guo_countering_2018} and we found three Input Alteration defences for DRL \cite{ohashi_deep_2021,guo_edge_2021,gleave_adversarial_2020}. These methods have a risk of removing valuable data from the input that could improve the performance of an agent as shown by the observation masking defence \cite{gleave_adversarial_2020}. However, using an auto-encoder did improve the performance over the original agent \cite{ohashi_deep_2021}.

Methods that detect adversarial inputs are able to more precisely blind the agent to those inputs or sanitise the observation to remove the malicious components. Message filtering was used against both perturbed \cite{xue_mis-spoke_2022} and malicious communications\cite{mitchell_gaussian_2020}. However, attacks have been shown to avoid detection \cite{carlini_adversarial_2017} in a supervised learning context. We believe that this evasion of detection has yet to be considered in the context of RL.


\noindent \textbf{Adversarial Detection}
identifies either the existence or location of adversarial attacks. Tekgul et al. \cite{tekgul_real-time_2022} considers adversarial detection but does not consider how to remove or repair adversarial observations.

\begin{table*}[ht]
\centering
\caption{Input Alteration Defences}
\begin{tabular}{p{3.6cm}p{1.8cm}p{2.6cm}p{1.8cm}p{2.4cm}l} 
\hline
\textbf{Name of defence}                          & \textbf{Requires Adversarial Detection} & \textbf{Counters attacks}  & \textbf{Impact on performance} & \textbf{Defended algorithm} & \textbf{Framework}  \\ 
\hline
\textbf{DQWAE \cite{ohashi_deep_2021}}                                    & $\times$                                  & Observation Perturbation    & Positive                       & DQN                        & MDP             \\ 
\textbf{Blinding \cite{guo_edge_2021}}                                     & $\times$                                  & Natural Adversarial Examples & Not evaluated                  & EDGE                       & MDP             \\ 
\textbf{Message Filtering for Robust Cooperation \cite{mitchell_gaussian_2020}} & $\checkmark$                                 & Malicious Communications    & Not evaluated                  & Cooperative Agents         & DEC-POMDP       \\ 
\textbf{Instance-based defense \cite{garcia_instance-based_2022}}                   & $\checkmark$                                 & Observation Perturbation    & Not evaluated                  & DQN                        & MDP             \\ 
\textbf{Message filter \cite{xue_mis-spoke_2022}}                           & $\checkmark$                                 & Communication Pertrubations  & Not evaluated                  & CommNet, TarMAC, NDQ       & Dec-POMDP-Com   \\
\textbf{Observation masking \cite{gleave_adversarial_2020}}                           & $\times$                                 & Natural Adversarial Examples  & Negative       & PPO      & SG   \\
\hline
\end{tabular}
\label{tab:input_alt}
\end{table*}



\noindent \textbf{Regularisation}
is a method of avoiding overfitting and has been used as a defence against AML attacks through the regularisation of the Lipschitz Constant \cite{tang_deep_2019}.

\noindent \textbf{Memory}
allows an agent to operate in a partially observable environment. Deep Recurrent Q-Networks use a Long Short-Term Memory component to remember a latent state representation \cite{hausknecht_deep_2015}. This is considered as a potential AML defence in \cite{russo_optimal_2019}, however we believe that further research is required into the potential risks of this defence. An adversary could target the memory component of the algorithm to increase the effectiveness of a single-step attack to impact future steps.

\section{Frameworks}

We discuss the various frameworks that are currently being used to model AML attacks against DRL and MARL. We then identify a number of challenges in these frameworks, and propose two new frameworks that address them.

\subsection{Reinforcement Learning Frameworks}
\label{sec:rl_model}

Agents in reinforcement learning \cite{kaelbling_reinforcement_1996}  interact with an environment to optimise return. The environment is described by a framework, e.g., a Markov Decision Process (MDP) \cite{bellman_markovian_1957}, Partially Observable Markov Decision Process (POMDP) \cite{astrom_optimal_1965}, Decentralised Partially Observable Markov Decision Process (DEC-POMDP) \cite{bernstein_complexity_2013}, Stochastic Game (SG) \cite{shapley_stochastic_1953}, or a Partially Observable Stochastic Game (POSG) \cite{hansen_dynamic_2004}.  Table \ref{tab:models} captures the suitability of these frameworks in modelling our identified attack vectors.  

\begin{table*}[ht]
\centering
\caption{Reinforcement learning frameworks and their ability to model different AML Attack Vectors}
\begin{tabulary}{1.0\textwidth}{lccccc}
\hline
\multirow{2}{*}{\textbf{Framework}}                   & \textbf{Observation} & \textbf{Action}   & \textbf{Communication}   & \textbf{Malicious}   & \textbf{Natural Adversarial} \\
&\textbf{Perturbations} & \textbf{Perturbations} & \textbf{Perturbations} & \textbf{Communications} & \textbf{Examples} \\ \hline
\textbf{MDP \cite{bellman_markovian_1957}}                                                                                           & $\times$                       & $\times$                   & $\times$                          & $\times$                       & $\times$                           \\ 
\textbf{POMDP \cite{astrom_optimal_1965}}                                                                                         & $\times$                        & $\times$                   & $\times$                          & $\times$                       & $\times$                           \\ 
\textbf{DEC-POMDP \cite{bernstein_complexity_2013}}                                                                                     & $\times$                        & $\times$                   & $\times$                          & $\times$                       & $\times$                           \\ 
\textbf{SG \cite{shapley_stochastic_1953}}                                                                               & $\times$                        & $\times$                   & $\times$                          & $\checkmark$                      & $\checkmark$                          \\ 
\textbf{POSG \cite{hansen_dynamic_2004}}                                                                                          & $\times$                        & $\times$                   & $\times$                          & $\checkmark$                      & $\checkmark$                          \\ 
\textbf{PR-MDP \cite{tessler_action_2019}}                                                    & $\times$                        & $\checkmark$                  & $\times$                         & $\times$                       & $\times$                           \\ 
\textbf{NR-MDP \cite{tessler_action_2019}}                                                            & $\times$                        & $\checkmark$                  & $\times$                         & $\times$                       & $\times$                           \\ 
\textbf{SA-MDP \cite{zhang_robust_2020-1}}                                                                                        & $\checkmark$                       & $\times$                   & $\times$                          & $\times$                       & $\times$                           \\ 
\hline
\end{tabulary}

\label{tab:models}
\end{table*}

In a MDP \cite{bellman_markovian_1957} (in Definition \ref{def:mdp}), a single agent is operating in an environment where it is able to observe the state of the entire environment. At each time step, the agent selects a single action from the action space, causing the state to probabilistically transition to a new state. MDPs do not feature an adversary and so are unable to model any AML Attack Vector. Despite this, this framework is the most commonly used for AML in DRL literature \cite{ying_towards_2021,kim_trc_2022,inkawhich_snooping_2020,chen_understanding_2021,zheng_vulnerability_2021,behzadan_vulnerability_2017,behzadan_whatever_2017,sun_who_2021,wang_adversarial_2020,korkmaz_daylight_2020,weng_toward_2020,chan_adversarial_2020,wu_adversarial_2021,qu_adversary_2021,qu_attacking_2021,everett_certifiable_2021,hussenot_copycat_2020,praveen_kumar_critical_2021,wu_crop_2022,ohashi_deep_2021,guo_edge_2021,garcia_instance-based_2022,huai_malicious_2020,lee_query-based_2021,oikarinen_robust_2021,tretschk_sequential_2018,lee_spatiotemporally_2020,sun_stealthy_2020,lin_tactics_2017,buddareddygari_targeted_2021}.

\begin{definition}
A Markov Decision Process (MDP) is defined by the 4-tuple $(S, A, T, R)$, in which $S$ is the set of states, $A$ is the set of actions, $T$ is the state transition function that stochastically maps a state $s \in S$ and action $a \in A$ to a next state $s' \in S$ such that $T(s, a) = s'$, and $R$ is the reward function.
\label{def:mdp}
\end{definition}

A POMDP \cite{astrom_optimal_1965} (see  Definition \ref{def:pomdp}) is an extension of an MDP that separates the state from the observation. In doing so, the agent observes only a part of the state and must infer the true state from partial observations. The addition of partial observability does not improve the ability of the POMDP to represent different AML Attack Vectors because a POMDP does not allow for the modelling of an adversary.

\begin{definition}
A Partially Observable Markov Decision Process (POMDP) is defined by the 6-tuple $(S, A, T, \Omega, O, R)$, in which $S$ is the set of states, $A$ is the set of actions, $T$ is the state transition function that maps a state $s \in S$ and action $a \in A$ to a next state $s' \in S$ such that $s' = T(s, a)$, $\Omega$ is the set of observations, $O$ is the observation probability function that maps a state $s \in S$ and action $a \in A$ to an observation $o \in \Omega$ such that $O(s, a) = o$, and $R$ is the reward function.
\label{def:pomdp}
\end{definition}

A DEC-POMDP \cite{bernstein_complexity_2013} (see Definition \ref{def:dec_pomdp}) extends a POMDP to cooperative multi-agent environments. Multiple agents act simultaneously in this framework and all agents receive the same reward \cite{oliehoek_concise_2016}. The state transition, reward, and observation functions are dependent on the joint actions of all agents in the environment. This framework is used in AML for MARL \cite{blumenkamp_emergence_2021,lin_robustness_2020}. However, DEC-POMDPs cannot represent any of the AML Attack Vectors that we have identified because of the fully cooperative nature of DEC-POMDPs. The lack of a competitive reward signal prevents the existence of an adversarial agent which may use an attack vector to affect other agents in the environment.

\begin{definition}
A Decentralised Partially Observable Markov Decision Process (DEC-POMDP) is defined by the 7-tuple $(I, S, \{A_i\}, T, \{\Omega_i\}, O, R)$, in which $I$ is the set of agents, $S$ is the set of states, $\{A_i\}$ is the joint set of action sets, $A_i$ is the action set of agent $i \in I$, $T$ is the state transition function that maps a state $s \in S$ and joint action $\{a_i\} \in \{A_i\}$ to a next state $s' \in S$ such that $s' = T(s, \{a_i\})$, $\{\Omega_i\}$ is the joint set of observations, $\Omega_i$ is the set of observations of agent $i \in I$, $O$ is the joint observation probability function that maps a state $s \in S$ and joint action $\{a_i\} \in \{A_i\}$ to a joint observation $\{o_i\} \in \{\Omega_i\}$ such that $O(s, \{a_i\}) = \{o_i\}$, and $R$ is the reward function.
\label{def:dec_pomdp}
\end{definition}

A SG \cite{shapley_stochastic_1953} (Definition \ref{def:sg}) is focused on multi-agent environments. This framework provides each agent with an independent reward thus allowing competitive, cooperative and mixed multi-agent environments. These environments are able to contain adversaries, providing the framework with the capability of modelling the Natural Adversarial Examples attack vector \cite{pinto_robust_2017,guo_adversarial_2021,gleave_adversarial_2020}, since other agents take actions that affect the transition of the state. Furthermore, the joint action set may include communication actions that other agents  use to perform an AML attack. This allows SGs to represent the Malicious Communications attack vector.

\begin{definition}
A Stochastic Game (SG) is defined by the 5-tuple $(I, S, \{A_i\}, T, \{R_i\})$, in which $I$ is the set of agents, $S$ is the set of states, $\{A_i\}$ is the joint set of action sets, $A_i$ is the action set of agent $i \in I$, $T$ is the state transition function that maps a state $s \in S$ and joint action $\{a_i\} \in \{A_i\}$ to a next state $s' \in S$ such that $s' = T(s, \{a_i\})$, $\{R_i\}$ is the set of reward functions, and $R_i$ is the reward function of agent $i \in I$.
\label{def:sg}
\end{definition}

A POSG  \cite{hansen_dynamic_2004} (Definition \ref{def:posg}) extends a SG to separate agents' observations from the state similar to what POMDPs do for MDPs. Like the SG, a POSG can model Malicious Communications and Natural Adversarial Examples \cite{phan_learning_2020}.

\begin{definition}
A Partially Observable Stochastic Game (POSG) is defined by the 7-tuple $(I, S, \{A_i\}, T, \{\Omega_i\}, O, \{R_i\})$, $I$ is the set of agents, $S$ is the set of states, $\{A_i\}$ is the joint set of action sets, $A_i$ is the action set of agent $i \in I$, $T$ is the state transition function that maps a state $s \in S$ and joint action $\{a_i\} \in \{A_i\}$ to a next state $s' \in S$, $s' = T(s, a)$, $\{\Omega_i\}$ is the joint set of observations $\Omega_i$ for agent $i \in I$, $O$ is the joint observation probability function that maps a state $s \in S$ and joint action $\{a_i\} \in \{A_i\}$ to a joint observation $\{o_i\} \in \{\Omega_i\}$ such that $O(s, \{a_i\}) = \{o_i\}$, $\{R_i\}$ is the set of reward functions $R_i$ for agent $i \in I$.
\label{def:posg}Probabilistic Action Robust
MDP

\end{definition}

\subsection{AML for DRL Frameworks}
\label{sec:aml_model}
Many  papers that consider AML attacks against DRL use standard frameworks. However,  some  consider variations of these frameworks to better describe  interactions between AML attacks and the targeted agents, such as  the State Adversarial MDP (SA-MDP) \cite{zhang_robust_2020-1}, the Probabilistic Action Robust MDP (PR-MDP) \cite{tessler_action_2019}, and the Noisy Action Robust MDP (NR-MDP) \cite{tessler_action_2019}.

The PR-MDP \cite{tessler_action_2019} (Definition \ref{def:pr_mdp}) adds an adversary to the standard MDP that is probabilistically able to take an action in place of the original agent. This framework represents Action Perturbations, however as it is single agent and unable to represent adversarial changes to the observation, it cannot represent any other AML attack vector.

\begin{definition}
A Probabilistic Action Robust MDP (PR-MDP) is defined by the 6-tuple $(S, A, T, R, v, \alpha)$, in which $S$ is the set of states, $A$ is the set of actions, $T$ is the state transition function that maps a state $s \in S$ and action $a \in A$ to a next state $s' \in S$ such that $T(s, a) = s'$, $R$ is the reward function, and $v$ is the Probabilistically-Robust (PR) operator such that $v(s) = a'$ where $a' \in A$ is either an adversarial action with probability $\alpha$ or the agent's original action with probability $1-\alpha$ that is executed in the state transition function.
\label{def:pr_mdp}
\end{definition}

The NR-MDP \cite{tessler_action_2019} (see Definition Definition \ref{def:nr_mdp}) adds an adversary to the standard MDP that is able to add a perturbation into the action selection of an agent. Like PR-MDPs, NR-MDPs only represent Action Perturbation Attack Vectors.

\begin{definition}
A Noisy Action Robust MDP (NR-MDP) is defined by the 6-tuple $(S, A, T, R, v, \Delta)$, in which $S$ is the set of states, $A$ is the set of actions, $T$ is the state transition function that maps a state $s \in S$ and action $a' \in A$ to a next state $s' \in S$ such that $T(s, a') = s'$, $R$ is the reward function, and $v$ is the noisy-robust operator such that $v(s, a) = a'$ where $a, a' \in A$ and $|a - a'| < \Delta$, $a'$ is the perturbed action that is executed in the state transition function, and $\Delta$ is the magnitude of the perturbation.
\label{def:nr_mdp}
\end{definition}

The SA-MDP \cite{zhang_robust_2020-1} (Definition \ref{def:sa_mdp}) adds an adversary to an MDP that is able to alter the state observation that the agent usually receives. The SA-MDP is capable of representing perturbations to the observation through the adversarial state-permutation function and is used as such in the literature \cite{korkmaz_adversarial_2021,korkmaz_assessing_2021,korkmaz_investigating_2021,korkmaz_non-robust_2021,tekgul_real-time_2022,zhang_robust_2020-1,qiaoben_understanding_2021}. The SA-MDP cannot represent any other AML attack vector due to the restriction on adversaries present in MDPs.

\begin{definition}
A State-Adversarial Markov Decision Process (SA-MDP) is defined by the 5-tuple $(S, A, T, R, v)$, in which $S$ is the set of states, $A$ is the set of actions, $T$ is the state transition function that maps a state $s \in S$ and action $a \in A$ to a next state $s' \in S$ such that $T(s, a) = s'$, $R$ is the reward function, and $v$ is the adversarial state-permutation function such that $v(s) = s'$ where $s, s' \in S$, $s$ is the original state and $s'$ is the state as observed by the agent. This state permutation is only a change to the perception of the agent and does not change the true state of the environment.
\label{def:sa_mdp}
\end{definition}

\subsection{Proposed AML for MARL Frameworks}

POMDP and DEC-POMDP are suitable frameworks for modelling single-agent and cooperative multi-agent reinforcement learning respectively. However, they are unable to model Malicious Communications or Natural Adversarial Examples. On the other hand, POSGs are able to effectively model both of these Attack Vectors.
We thus propose two new frameworks that extend and combine the existing frameworks presented in sections \ref{sec:rl_model} and \ref{sec:aml_model}, namely the Observation Adversarial POSG (OA-POSG) (Definition \ref{def:oa_posg}) and Action Robust POSG (AR-POSG) (Definition \ref{def:ar_posg}).

\textit{OA-POSG} (Definition \ref{def:oa_posg}) is the extension of the POSG with adversarially perturbed observations, allowing Observation Perturbations to be applied to any form of multi-agent environment. We introduce an observation-adversarial function, $v_i$, for each agent $i$ in the environment. This function perturbs the observation of an agent, similar to the single-agent SA-MDP \cite{zhang_robust_2020-1}. We introduce a parameter that captures the magnitude of perturbation, $\Delta_i$. This parameter allows us to model the magnitude of change that an adversary may inflict on the observation. OA-POSG features a scope function, $\Sigma_i$, which captures what elements of an observation are perturbed. The tempo function $\Theta$, in the OA-POSG, determines if a particular state will be attacked and captures the concept of when an attack will occur.

\begin{definition}
An Observation-Adversarial Partially Observable Stochastic Game (OA-POSG) is defined by the 11-tuple $(I, S, \{A_i\}, T, \{\Omega_i\}, O, \{R_i\}, \{v_i\}, \{\Theta_i\}, \{\Delta_i\}, \{\Sigma_i\})$
\begin{itemize}
    \item $I$ is the set of agents. The subscript $i$ indicates that parameter or function is unique to the $i$th agent
    \item $S$ is the set of states
    \item $\{A_i\}$ is the joint action space of all agents
    \item $T$ is the state transition function that maps a state $s \in S$ and joint action $\{a_i\} \in \{A_i\}$ to a next state $s' \in S$ such that $T(s, a) = s'$
    \item $\{\Omega_i\}$ is the joint set of observations $\Omega_i$
    \item $O$ is the joint observation probability function
    \item $\{R_i\}$ is the set of reward functions for all agents
    \item $\Theta_i$ is the tempo of attack determines if the observation of the agent $i$ for a given state $s$ will be perturbed by the function $v_i$
    \item $\Delta_i$ is the maximum magnitude of the change of a single element in the perturbed observation  $e_{ij} \in o_i'$ from the original observation $o_i$ of agent $i$
    \item $\Sigma_i$ is the scope of the attack which determines which elements of the observation $o_i$ is perturbed at state $s$. $\Sigma_i(s) = \{e_{ij}\}$, where element $e_{ij} \in o_i$ will be perturbed by the function $v_i$ and any element not in the returned set will not be changed
    \item $\{v_i\}$ is the set of adversarial observation-permutation functions, such that the observation received by agent $i$ is
    
    $o_i' = \begin{array}{lr}
         v_i(o_i, \Delta_i, \{e_{ij}\}) & \textit{if } \Theta_i(s) \leq \alpha \\
         o_i & \textit{if } \Theta_i(s) > \alpha
    \end{array}$
\end{itemize}
\label{def:oa_posg}
\end{definition}

We have not found a requirement for a framework that models both input and output perturbations so we do not consider any extensions that would combine either the PR-MDP or NR-MDP with the OA-POSG. The goal of Observation Perturbations is to affect the actions selected by an agent, however Action Perturbations are able to directly alter these actions. Instead, we propose the AR-POSG which combines and extends the PR-MDP, NR-MDP, and POSG to allow the modelling of Action Perturbations in multi-agent environments. 

\textit{AR-POSG} (Definition \ref{def:ar_posg}) features an action-robust operator $v_i$ that adversarially perturbs the action of agent $i$. This allows the action of each agent in the environment to be subject to different perturbations. The action-robust operator is conditioned by $\Theta_i$ and $\Delta_i$. $\Theta_i$ is the attack tempo and determines whether action $a_i$ is perturbed at state $s$. $\Delta_i$ sets a maximum magnitude on the perturbation. 

\begin{definition}
An Action-Robust Partially Observable Stochastic Game (AR-POSG) is defined by the 10-tuple $(I, S, \{A_i\}, T, \{\Omega_i\}, O, \{R_i\}, \{v_i\}, \{\Theta_i\}, \{\Delta_i\})$,
\begin{itemize}
    \item $I$ is the set of agents. The subscript $i$ indicates that parameter or function is unique to the $i$th agent
    \item $S$ is the set of states
    \item $\{A_i\}$ is the joint action space of all agents
    \item $T$ is the state transition function that maps a state $s \in S$ and joint action $\{a_i\} \in \{A_i\}$ to a next state $s' \in S$ such that $T(s, a) = s'$
    \item $\{\Omega_i\}$ is the joint set of observations $\Omega_i$
    \item $O$ is the joint observation probability function
    \item $\{R_i\}$ is the set of reward functions for all agents
    \item $\Theta_i$ is the tempo of attack determines if the action of the agent $i$ for a given state $s$ will be perturbed by the function $v_i$
    \item $\Delta_i$ is the maximum magnitude of change from the perturbed action $a'$ to the original action $a_i$ where $ a, a' \in A_i$ of agent $i$
    \item $\{v_i\}$ is the set of action-robust operators, such that the action performed by agent $i$ is
    
    $a_i' = \begin{array}{lr}
         v_i(o_i, a_i, \Delta_i) & \textit{if } \Theta_i(s) \leq \alpha \\
         a_i & \textit{if } \Theta_i(s) > \alpha
    \end{array}$. 
    
    where $a_i \in A_i$ is the original action selected by the agent.
\end{itemize}
\label{def:ar_posg}
\end{definition}

The new frameworks of AR-POSG and OA-POSG enable more sophisticated modelling of execution-time AML attacks against both DRL and MARL. The frameworks allow us to model multiple simultaneous Attack Vectors and are the only frameworks capable of modelling Communication Perturbation attacks to the best of our knowledge. The frameworks enable the modelling of more detail in AML attacks, namely, the tempo, magnitude, and attack locations.


\section{Research Gaps and Challenges}
We focus  on  new AML attacks that may be effective against DRL and MARL algorithms, and suggest approaches to AML defences  that may improve  algorithm robustness.

\subsection{Attacks against Multiple Agents}
A major research direction that we have identified is AML attacks against multiple agents. The development and understanding of these attacks allow us to better understand the vulnerability and risks of using MARL. MARL uses both explicit \cite{foerster_learning_2016-1} and implicit communications \cite{gupta_cooperative_2017} to coordinate the agent behaviours and we believe that a cascading effect on the system may be produced by attacking a single agent. This failure induced by a single agent's actions is shown by works that present Malicious Communications \cite{tu_adversarial_2021, xue_mis-spoke_2022} and Natural Adversarial Example \cite{uesato_adversarial_2018, fujimoto_adversarial_2021, guo_adversarial_2021, pan_improving_2021, phan_learning_2020, gleave_adversarial_2020} attacks. Certain individuals in social networks are able to unduly influence the behaviour of the whole network \cite{yang_efficient_2016}. Likewise, an attack against an individual agent  may disproportionately affect the performance of the whole system. An attack may be more effective based on factors, such as timing and communication protocols. Measuring this impact is a very important research problem that could be used to improve robustness of multi-agent systems against attacks.

The influence of a single agent in a multi-agent system changes over the course of an episode. We believe that investigating the effect of switching the targeted agent of an attack is an important research direction. Findings may be used to increase the effectiveness of attacks or demonstrate flaws in AML defences that assume all agents are being attacked \cite{mitchell_gaussian_2020, xue_mis-spoke_2022}. Attack such as the Strategically Timed Attack \cite{lin_tactics_2017} and the Critical Point Attack \cite{sun_stealthy_2020} consider the importance of an action at a particular time step during a game. Extending these to consider the agent importance  at a particular time step would allow for more precise attacks against multi-agent systems.

Multi-agent systems may be \textit{homogeneous}, in which all agents share an identical policy, or  \textit{heterogeneous}, in which agents have different policies. The relative vulnerability to AML attacks of heterogeneous vs homogeneous systems has yet to be explored. We hypothesise that homogeneous systems are vulnerable based on the increased implicit communication \cite{gupta_cooperative_2017} and reduced variance between policies. 

Attacks that use reward-based targeting rely on the target agent attempting to optimise a single reward. A single reward is often used in cooperative MARL for all agents. However, mixed MARL can produce cooperation and coordination due to the use of certain reward functions \cite{jaques_social_2019}. We believe that it would be valuable to compare the vulnerability of MARL algorithms that use shared rewards to those that use individual rewards such as COMA \cite{foerster_counterfactual_2018}. 

\subsection{Attack Vectors}

The OA-POSG and AR-POSG frameworks allow research into combining different Attack Vectors that may produce more effective attacks than those that use a single attack vector. For example, allowing an adversarial agent to both play the opponent and discover Natural Adversarial Examples while simultaneously perturbing inter-agent communications could be much more effective than either of those attacks individually.

Research into the quantification of the impact of different Attack Vectors on different algorithms and environments would be highly valuable. We believe that attacks that give an adversary direct input to the target agents such as Observation and Communication Perturbations and Malicious Communications may be more effective than those that rely on indirect effects such as Natural Adversarial Examples.

The enumeration of Attack Vectors has revealed potential gaps in AML attacks against MARL, namely, white-box Malicious Communications, patch attacks, action-based and state-based targeted Communication Perturbation attacks, untargeted Communication Perturbation attacks, action-based and state-based targeted Natural Adversarial Examples. Malicious Communications allow an adversary to send any data to a target which we believe would be devastating as shown by the single-pixel attack in the supervised learning domain \cite{su_one_2019}. Similarly, the single-pixel attack could be investigated in the Patch Attack vector. Communication Perturbation has only been considered with reward-based targeting. However, no research has been done to our knowledge into targeting actions and states. Further, untargeted Communication Perturbation attacks have not been investigated. Natural Adversarial Examples have been used to target rewards, but no research has been done to see if natural examples exist that could manipulate a target into taking certain actions or moving towards certain states. Being able to convince an adversary to move towards specific states, or take certain actions could have a significant impact on the field of competitive MARL.

We did not find any requirement for a framework capable of modelling simultaneous Observation Perturbations and Action Perturbations. However, an advantage of investigating simultaneous Observation Perturbations and Action Perturbations may be to find a combined attack that can use a lower magnitude perturbation for both attacks.

\subsection{Quantifying Defence Generalisations}
Measuring the effectiveness of a defence against multiple different attack types would provide insight into practical AML mitigations. Our survey has shown that many AML defences are only evaluated against the attack they were designed to mitigate. DRL trained algorithms are already being used in vital systems and will be subject to a broad range of attacks. AML defences must be able to be deployed to protect these algorithms and it is vitally important that the impact of AML defences be well understood before deployment or risk increasing the overall system vulnerability. We believe that future work can contribute by evaluating the impact of any new defences against AML attacks on the performance of the unattacked algorithm.

Many defences focus on the training before an attack has occurred. An alternative defences could occur during an attack by identifying and removing adversarial examples. An ideal defence would selectively sanitise an observation to remove adversarial aspects of the observation but retain other important and unattacked information. We believe that using defences before and during an attack will provide better security than either option alone.

We found a single defence against Action Perturbation \cite{tessler_action_2019}. However these defences hold the potential to address both the mitigation of Action Perturbation attacks and to explore resilience in the face of attacks from other vectors. That is if a bad action occurs then the agent should learn how to best recover. This is a vital property for any system that may be operating in the real world. 

\subsection{Employing Attacker Metrics in Defence}

In cyber-security, intelligence about the adversary is vital to crafting an appropriate response. To this end, we believe that both identifying the presence of an attack and the properties of the attacker such as preferred Attack Vectors, tempo of attacks, etc. may produce more effective defences. For example, if an attack is periodic, that information can be used to better identify and remove future attacks.

Another key research direction is using the results of simulated attacks to reduce vulnerabilities. For example, if an attack consistently targets a few key state-action pairs to achieve its goal then more focus should be on those state-actions to reduce the vulnerability in those states.


\section{Conclusion}

There are significant gaps in the research around AML attacks and defences for MARL that need to be addressed, including mitigating AML attacks against multiple agents, the combined effect of multiple AML attacks, quantifying the effectiveness of AML defences, and using knowledge about an attacker to improve AML defences.
Our survey has identified numerous Adversarial Machine Learning (AML) attacks and defences for single-agent Deep Reinforcement Learning (DRL) algorithms. However we found few attacks and defences for Multi-Agent Reinforcement Learning (MARL). Beyond the much needed systematisation of knowledge, a key contribution of this work is the perspective of Attack Vectors used to understand attacks by capturing the means by which an adversary might execute an attack, i.e., action, observation, communication perturbations, malicious communications, and natural adversarial examples. Another key contribution of this work is the proposal of two new frameworks, Observation Adversarial Partially Observable Stochastic Game (OA-POSG) and Action Robust Partially Observable Stochastic Game (AR-POSG), which address a gap in the current frameworks being used to research AML in DRL, namely, the inability to model Communication Perturbations, the inability to model multiple simultaneous AML attack vectors, and the lack of detail around the tempo, magnitude and location of attacks. Future work identifies the need for studying attacks against multiple agents, and on quantifying the effectiveness of defences through the use of various metrics.

\newpage
\bibliographystyle{custom_ieeetr}
\bibliography{2022-08-18}

\appendix{}

In Table \ref{tab:evasion_att}, we present a list of Observation Perturbation attacks focusing on the attack information, the attack objective, whether the  attack is transferred, the algorithms that are attacked, and the framework used by the paper that presented the attack. 

Observation Perturbation attacks require carefully crafted alterations to induce an adversarial effect. AML attacks against supervised learning algorithms may use the Fast Gradient Sign Method (FGSM) \cite{goodfellow_explaining_2015} or Projected Gradient Descent (PGD) \cite{madry_towards_2018}, and are commonly used to craft Observation Perturbation attacks \cite{fujimoto_adversarial_2021,tu_adversarial_2021,huang_adversarial_2017,praveen_kumar_critical_2021,kos_delving_2017,qiaoben_understanding_2021}. Other methods find universal perturbations \cite{hussenot_copycat_2020,huai_malicious_2020,tekgul_real-time_2022}. However, some algorithms have been presented to focus on the unique characteristics of DRL. Chan et al. \cite{chan_adversarial_2020} used a Static Reward Impact Map to generate an attack by evaluating the impact that perturbations of different features had on the cumulative reward of an agent. Other methods use DRL to craft observation perturbations \cite{russo_optimal_2019,tretschk_sequential_2018,sun_stealthy_2020,qiaoben_strategically-timed_2021}. The other set of methods use the the gradient of the Q-value to craft an attack \cite{praveen_kumar_critical_2021,zhang_robust_2020-1,pattanaik_robust_2018}. 

The timing of an attack has been shown to be a key factor in the success of that attack \cite{sun_stealthy_2020, huang_adversarial_2017, lin_tactics_2017, qiaoben_strategically-timed_2021}. The simplest timing is the uniform attack \cite{huang_adversarial_2017}, which attacks at every time step. More efficient attacks wait until the value of a state or action is above a certain threshold \cite{lin_tactics_2017, qiaoben_strategically-timed_2021}. The number of time steps can further be decreased by considering the future effects of an attack \cite{sun_stealthy_2020}.


\begin{table*}
\caption{Observation Perturbation Attacks}
\begin{tabulary}{1.0\textwidth}{Lp{2cm}llLl} 
\hline
\textbf{Name of attack}                                                 & \textbf{Information for attack} & \makecell[lt]{\textbf{Objective} \\ \textbf{of attack}}   & \makecell[lt]{\textbf{Transfer} \\ \textbf{attack}} & \textbf{Target algorithm(s)} & \textbf{Framework}  \\ 
\hline
\textbf{SRIMA \cite{chan_adversarial_2020}}                                                          & Black-box                       & Reward-based         & $\checkmark$                       & DQN, PPO                   & MDP    \\
\textbf{Adversarial Examples \cite{huang_adversarial_2017}}                                 & Black-box                       & Reward-based         & $\checkmark$                      & A3C, TRPO, DQN         & None             \\ 
\textbf{CopyCAT \cite{hussenot_copycat_2020}}                                              & Black-box                       & Action           & $\checkmark$                      & DQN                        & MDP             \\ 
\textbf{Critical State Detection \cite{praveen_kumar_critical_2021}}                                       & White-box                       & Reward-based         & $\times$                       & A3C                        & MDP             \\ 
\textbf{VF \cite{kos_delving_2017}}                                                             & White-box                       & Untargeted         & $\times$                       & A3C                        & None             \\ 
\textbf{MO-RL attack \cite{garcia_learning_2020}}                                                   & Black-box                       & Reward-based         & $\times$                       & MODQN                        & MDP             \\ 
\textbf{Optimal Attack  \cite{russo_optimal_2019}}                                                 & Black-box                       & Reward-based         & $\times$                       & DDPG, DRQN            & POMDP           \\ 
\textbf{OSFW(U) \cite{tekgul_real-time_2022}}                                          & White-box                       & Untargeted                  & $\times$                       & DQN, PPO, A3C              & SA-MDP             \\ 
\textbf{UAP \cite{tekgul_real-time_2022}}                                          & White-box                       & Untargeted                  & $\times$                       & DQN, PPO, A3C              & SA-MDP             \\ 
\textbf{RS \cite{zhang_robust_2020-1}}             & White-box                       & Reward-based         & $\times$                      & PPO, DDPG, DQN             & SA-MDP             \\ 
\textbf{MAD \cite{zhang_robust_2020-1}}             & White-box                       & Reward-based         & $\times$                      & PPO, DDPG, DQN             & SA-MDP             \\ 
\textbf{Naive Adversarial attack \cite{pattanaik_robust_2018}}                                       & Black-box                       & Untargeted                  & $\times$                       & DDQN, DDPG                 & None             \\ 
\textbf{Gradient based attack \cite{pattanaik_robust_2018}}                                       & White-box                       & Reward-based                  & $\times$                       & DDQN, DDPG                 & None             \\ 
\textbf{ATN \cite{tretschk_sequential_2018}}                          & White-box                       & Reward-based & $\times$                       & DQN                      & MDP             \\ 
\textbf{Snooping attack \cite{inkawhich_snooping_2020}}                            & Black-box                       & Untargeted                  & $\checkmark$                      & DQN, PPO                   & MDP             \\ 
\textbf{Antagonist Attack \cite{sun_stealthy_2020}}                       & White-box                       & Reward-based         & $\times$                       & A3C, DDPG, PPO             & MDP             \\ 
\textbf{Critical Point Attack \cite{sun_stealthy_2020}}                       & White-box                       & Reward-based         & $\times$                       & A3C, DDPG, PPO             & MDP             \\ 
\textbf{Tentative Frame Attack \cite{qiaoben_strategically-timed_2021}}                                         & White-box                       & Reward-based         & $\times$                       & PPO                        & SS-MDP             \\ 
\textbf{STA \cite{lin_tactics_2017}}                                     & White-box                       & Reward-based         & $\times$                       & DQN, A3C                   & MDP             \\ 
\textbf{EA \cite{lin_tactics_2017}}                                              & White-box                       & State-based                & $\times$                       & DQN, A3C                   & MDP             \\ 
\textbf{Two-stage attack \cite{qiaoben_understanding_2021}}                                               & White-box                       & Untargeted                  & $\times$                       & DQN, A2C, PPO              & SA-MDP          \\ 
\textbf{Grid Manipulation Attack \cite{chen_understanding_2021}}             & Black-box                       & State-based                & $\times$                       & D3QN, PPO, A2C             & MDP             \\ 
\textbf{Action Distortion Attack \cite{chen_understanding_2021}}             & Black-box                       & State-based                & $\times$                       & D3QN, PPO, A2C             & MDP             \\ 
\makecell[lt]{\textbf{Criticality-Based Adversarial} \\ \textbf{Perturbation \cite{zheng_vulnerability_2021}}}                     & White-box                       & Reward-based         & $\times$                       & DQN                        & MDP             \\ 
\textbf{Policy Induction Attack \cite{behzadan_vulnerability_2017}}                                        & Black-box                       & Untargeted                  & $\checkmark$                      & DQN                        & MDP             \\ 
\textbf{Contiguous attack \cite{behzadan_whatever_2017}}                                              & White-box                       & Untargeted                  & $\times$                       & DQN                        & MDP             \\ 
\textbf{PA-AD  \cite{sun_who_2021}}                      & White-box                       & Reward-based         & $\times$                       & DQN, A2C, PPO              & MDP             \\ 
\textbf{OWR \cite{lin_robustness_2020}} & White-box                       & Reward-based         & $\times$                       & QMIX                       & DEC-POMDP       \\ 
\textbf{FGSM-based attack \cite{wang_adversarial_2020}}                                                           & White-box                       & Untargeted                  & $\times$                       & DQN, IQN                   & MDP             \\ 
\textbf{FD Method\cite{wang_adversarial_2020}}                                                      & Black-box                       & Untargeted                  & $\times$                       & DQN, IQN                   & MDP             \\ 
\textbf{Model Based Attack \cite{weng_toward_2020}}                                             & White-box                       & Reward-based         & $\times$                       & D4PG                       & MDP             \\
\textbf{Attacks on Beliefs \cite{fujimoto_adversarial_2021}} & White-box & Reward-based & $\times$ & CAPI & PuB-MDP \\
\hline
\end{tabulary}
\label{tab:evasion_att}
\end{table*}
\end{document}